# A Large Dataset of Object Scans

Sungjoon Choi[1], Qian-Yi Zhou[2], Stephen Miller[3], and Vladlen Koltun[2]

[1]Google Inc.    [2]Intel Labs    [3]Fyusion Inc.

**Abstract.** We have created a dataset of more than ten thousand 3D scans of real objects. To create the dataset, we recruited 70 operators, equipped them with consumer-grade mobile 3D scanning setups, and paid them to scan objects in their environments. The operators scanned objects of their choosing, outside the laboratory and without direct supervision by computer vision professionals. The result is a large and diverse collection of object scans: from shoes, mugs, and toys to grand pianos, construction vehicles, and large outdoor sculptures. We worked with an attorney to ensure that data acquisition did not violate privacy constraints. The acquired data was placed irrevocably in the public domain and is available freely at http://redwood-data.org/3dscan.

## 1 Introduction and Related Work

The presented dataset provides over 10,000 dedicated 3D scans of individual objects. The scans were produced "in the wild" in conditions designed to simulate broad deployment of 3D scanning systems to consumers. The operators were not experts in computer vision and did not have direct familiarity with computer vision research. The scans were thus produced in conditions that may be similar to those encountered when consumer-grade RGB-D cameras are used for object reconstruction by a broad base of users.

Widely deployed object reconstruction systems are likely to be based on mobile consumer-grade cameras, operated outside controlled lab environments by users without extensive training or expertise. The operators may hold the camera or the object in their hands and move these freely during scanning. Our dataset was collected in this mode and may help understand the challenges faced by object reconstruction pipelines in conditions of broad deployment.

A number of datasets of object scans have been put together in the past. The dataset of Lai et al. [4] includes 300 objects recorded with a consumer-grade RGB-D camera. This dataset is geared towards object detection and recognition applications and was recorded in the lab, using a localized camera and a computer-controlled turntable. The B3DO dataset [2] contains 849 RGB-D images of objects in their natural environments. This dataset is likewise motivated by object detection and includes individual images rather than video sequences: this is generally inadequate for high-fidelity reconstruction of complete objects. The BigBIRD dataset [6] provides 3D scans of 100 objects, acquired in the lab using a computer-controlled photobench and a static calibrated imaging setup.

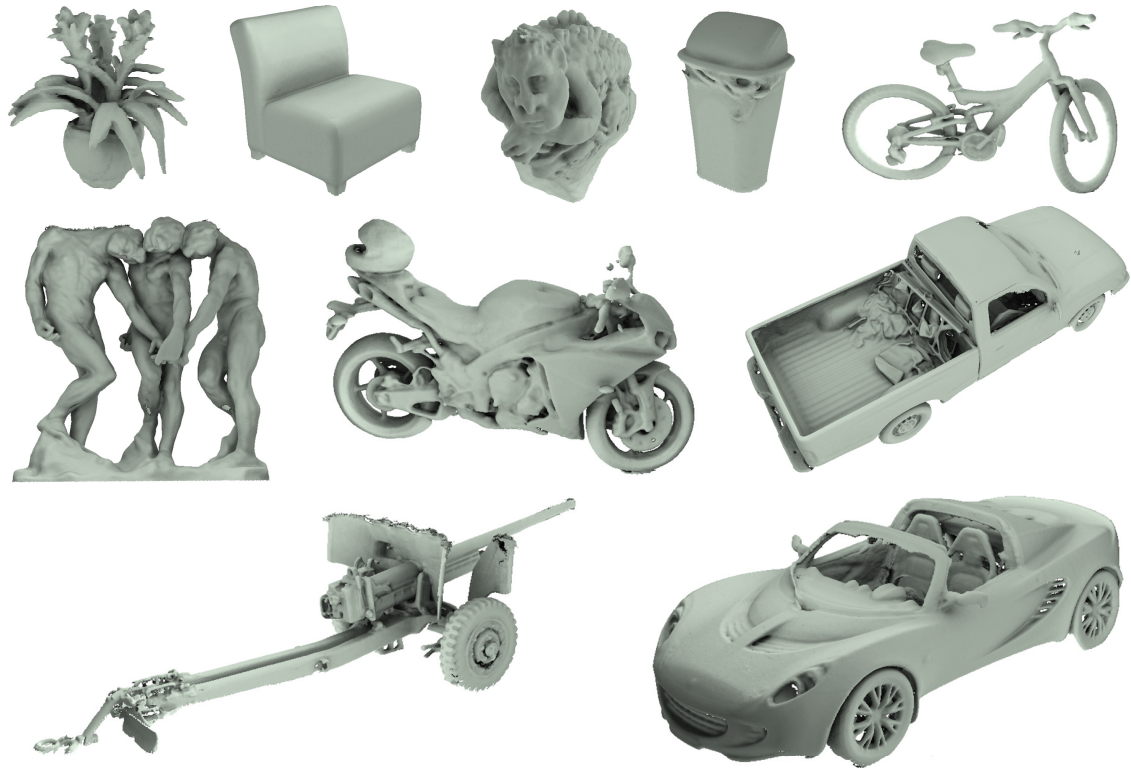

**Fig. 1.** A variety of objects reconstructed from our 3D scans.

Our dataset is one to two orders of magnitude larger than these prior collections and provides the kind of data that is likely to be encountered by broadly deployed object reconstruction systems.

## 2   Data Acquisition

To create the dataset, we put together ten lightweight mobile scanning setups. Each setup consists of a netbook, an RGB-D camera, and a lightweight carrying case. The netbooks were purchased off the shelf at large discount department stores. We used the Acer Aspire, a 1.4kg model with 4GB of RAM and an Intel Celeron 847 processor capable of streaming RGB-D data at 30 fps. A PrimeSense Carmine RGB-D camera was attached to the back of the netbook's display by a Velcro strip and powered off the netbook's USB port. The setup was thus fully mobile and could be operated as a single handheld unit, with a front-facing display and a back-facing RGB-D camera as shown in Figure 2(a). The camera could also be detached from the laptop and moved independently.

A custom scanning application was installed on each netbook. The application shows a live feed from the PrimeSense color camera. To begin recording, the operator simply presses the space bar. To help the operators keep the camera at a suitable distance from the scanned object, the live feed is color coded: everything that lies beyond 2 meters is colored red. The front-end of the scanning application is shown in Figure 2(b). To stop recording, the operator presses the space bar again. The recorded videos can be managed in a separate mode in the application.

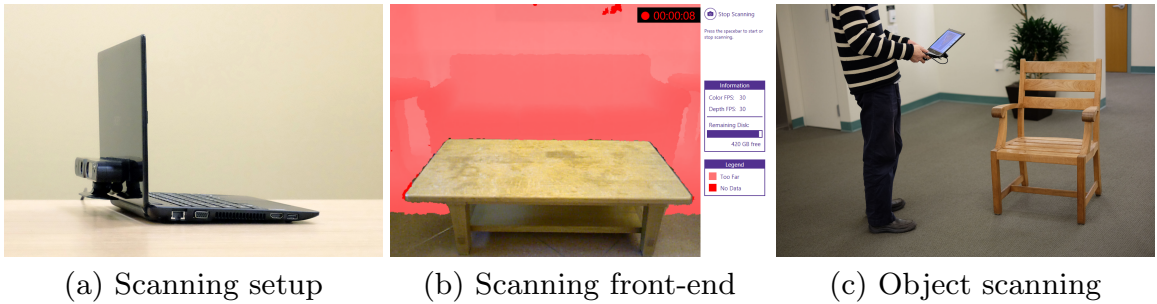

(a) Scanning setup      (b) Scanning front-end      (c) Object scanning

**Fig. 2.** (a) Mobile scanning setups provided to the operators. (b) Software front-end. (c) Demonstration of object scanning.

To recruit operators, we posted flyers across a large university campus and surrounding residential areas. Seventy operators were recruited in this manner, with up to ten working in parallel at any given time. The project was conducted with IRB approval. The operators were instructed to acquire videos of objects for the purpose of three-dimensional reconstruction: one video per object. Each operator watched a five-minute video tutorial and was given brief in-person instruction in the operation of the scanning setup. The tutorial and instruction demonstrated the use of the setup for scanning an object and asked the operator to perform a trial scan in our presence. We also went over a list of written guidelines that were handed out with the setup. We emphasized several aspects of scanning form: avoid direct sunlight, move slowly and smoothly, and cover the surface of the object thoroughly. We made sure that the operators understand that they are recording raw material for the creation of three-dimensional models of the objects they are scanning. The operators were thus guided to image as much of the objects' surface as possible. In particular, the operators were instructed to only scan objects that could be imaged clearly from at least three sides and from the top.

To encourage thorough and complete scans, the operators were compensated per minute of recorded data, rather than per object. We screened all videos submitted by the operators to prune out partial scans, repeated scans of the same object, and otherwise inadequate scans that did not meet the provided guidelines.

The choice of objects was left largely to the operators' discretion. A few restrictions were given due to the limitations of the camera. The guidelines instructed the operators to avoid objects that are too small or slender, such as pens or cords, objects that do not have a stable form, such as bedding and towels, and objects that are made entirely of transparent or highly reflective materials, such as mirrors or carafes. The written guidelines also provided a list of more than one hundred example objects. We emphasized that these are merely examples and that the list is far from exhaustive. In practice, the operators scanned many objects that were not on the list.

The tutorial video and our in-person instruction demonstrated two approaches to object scanning. The first approach is to keep the camera attached to the net-book and to move the scanning setup around a stationary object. The second approach is to arrange the camera and the netbook such that the operator can rotate a handheld object in front of the camera while monitoring the live feed on the netbook screen. Our general guideline was that if the operator can hold an object without difficulty in their hands, this object should be scanned in-hand, by turning it around in the field of view of the camera. Such scans will be referred to as "handheld." For larger or heavier objects, the operators were instructed to move the camera around the object, with the object itself kept stationary on a supporting surface such as the floor or a tabletop. Such scans will be referred to as "stationary."

To assist the scanning of tall objects, such as cargo vans and outdoor sculptures, we purchased a number of telescoping monopods and augmented them with mounts and Velcro strips. The camera could be attached to the monopod, which could be extended to a length of 1 meter, thus increasing the operator's reach. The monopods were provided to operators who were interested in scanning larger objects.

To make sure that the acquired data does not violate any restrictions, particularly with regards to privacy, we retained an attorney who composed suitable instructions and an agreement that was signed by the operators. As a result, all data in the dataset is fully compliant with the relevant laws in its original form, without post-hoc anonymization. The operators were fully aware that the data they are acquiring will be made public for research and development purposes. Neither they nor we nor our institutions retain any rights or ownership interests in the data. All rights were assigned to the public domain.

## 3    Dataset Composition

To assist the use of the dataset, we manually categorized the scans. Since our dataset is not intended for evaluation of recognition algorithms, some of our categories comprise semantically distinct object types. For example, the *personal grooming* category includes shampoo, soap, deodorant, and similar objects. Other categories are based on shape, such as *box* and *bottle*.

The "h-index" of the dataset is 44: there are 44 categories that have at least 44 scans each. The distribution of scans in these 44 categories is shown in Figure 3. The largest category, *car*, groups passenger vehicles such as sedans, SUVs, and vans. *Utility vehicles* are a separate category, as are *motorcycles* and *trucks*. Together, vehicles account for about 13% of the dataset. *Chair*, *table*, and *bench* together account for another 10%.

The average length of a scan is 77 seconds, yielding over 23 million RGB-D images in total.

We were surprised by the diversity of objects scanned by the operators. Some of these are mundane objects in our environment that rarely come to mind despite their ubiquity, such as fire hydrants (52 scans) and parking meters (16 scans). Many were unexpected, such as a large personal collection of rare musical instruments, hundreds of scans of sculptures and other artifacts in museums and galleries, and a Howitzer.

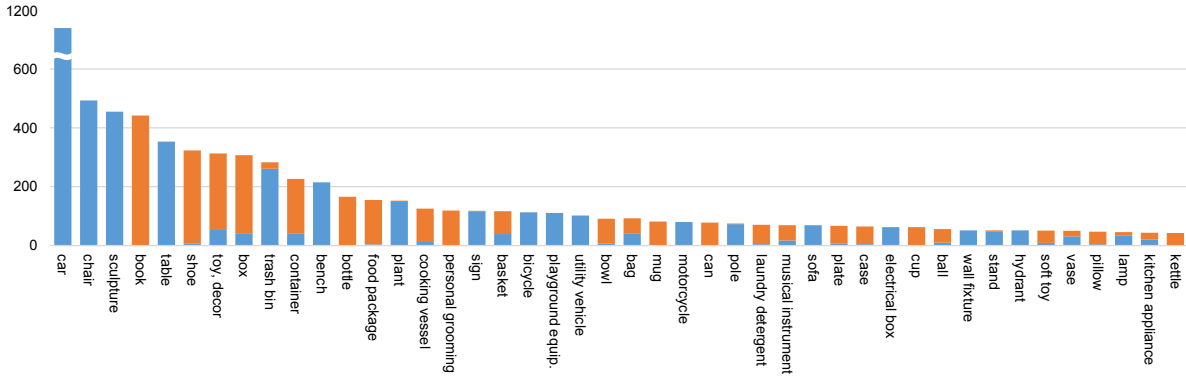

**Fig. 3.** Number of scans in the largest categories in the dataset. Blue for stationary scans, orange for handheld scans.

## 4 Reconstructed Models

Along with the raw RGB-D scans, we are releasing a number of reconstructed 3D models. The models were reconstructed by a pipeline that performs camera odometry and volumetric integration, building on the work of Newcombe et al. [5].

The ICP odometry used by Newcombe et al. is prone to drift and catastrophic failure in the presence of smooth surfaces. We have encountered this failure mode frequently in preliminary attempts to reconstruct scans from the collected dataset. For this reason, we have implemented a hybrid approach that combines ICP odometry with the RGB-D odometry of Kerl et al. [3], which aims to minimize the photometric error between consecutive RGB-D frames. The hybrid approach combines frame-to-model registration with the increased stability provided by the use of both geometric and photometric cues.

The hybrid odometry approach estimates transformation $\mathbf{T}_i$ for each frame $i$ by minimizing the following objective:

$$E(\mathbf{T}_i) = E_{\text{ICP}}(\mathbf{T}_i) + \lambda E_{\text{RGBD}}(\mathbf{T}_i). \tag{1}$$

$E_{\text{ICP}}(\mathbf{T}_i)$ is based on the point-to-plane error used in ICP registration:

$$E_{\text{ICP}}(\mathbf{T}_i) = \sum_{(\mathbf{p},\mathbf{q}) \in \mathcal{K}} \left\| (\mathbf{p} - \mathbf{T}_i \mathbf{q})^\top \mathbf{n_p} \right\|^2, \tag{2}$$

where $\mathbf{n_p}$ is the surface normal at $\mathbf{p}$ and $\mathcal{K}$ is a set of corresponding point pairs found by projective data association [5]. $E_{\text{RGBD}}(\mathbf{T}_i)$ is the photometric error between frames $i-1$ and $i$:

$$E_{\text{RGBD}}(\mathbf{T}_i) = \sum_{\mathbf{x}} \left\| I_i \left( \pi \left( \mathbf{T}_i^{-1} \mathbf{T}_{i-1} \pi^{-1}(\mathbf{x}, D_{i-1}) \right) \right) - I_{i-1}(\mathbf{x}) \right\|^2. \tag{3}$$

Here $D_k$ is the depth image from frame $k$, $\mathbf{x}$ iterates over pixels in frame $i-1$, $\pi(\cdot)$ is a projection operator that projects a 3D point in the camera coordinate frame to the image domain, $\pi^{-1}(\cdot)$ is an operator that produces a 3D point in the camera coordinate frame that corresponds to a given pixel, and $I_k(\mathbf{x})$ is the intensity of coordinate $\mathbf{x}$ in the color image from frame $k$ [3]. The weight $\lambda$ in Equation 1 balances the two error terms. It is chosen empirically and is identical for all reconstructions.

This pipeline does not handle loop closure and is not suitable for reconstructing large objects. We thus did not use it to reconstruct vehicles or sculptures. We also did not reconstruct handheld scans, because no pipelines we have tested performed satisfactorily on this data.

Objects from nine categories were reconstructed: chairs, tables, trash containers, benches, plants, signs, bicycles, motorcycles, and sofas. 1,781 sequences in these categories were processed. The pipeline lost track and failed to produce models for 969 sequences due to fast camera motion or odometry drift. The other 812 reconstructed models were qualitatively inspected and low-quality reconstructions were removed. 398 models passed the qualitative inspection and are available alongside all of the raw scans on the dataset web site. Table 1 lists the success rate of the reconstruction pipeline for each category.

Table 1. Reconstructed 3D models.

| Category | # sequences | # selected | Success rate |
|---|---|---|---|
| Chair | 465 | 125 | 27% |
| Table | 330 | 52 | 16% |
| Trash container | 263 | 73 | 28% |
| Bench | 211 | 21 | 10% |
| Plant | 146 | 41 | 28% |
| Sign | 112 | 18 | 16% |
| Bicycle | 109 | 22 | 20% |
| Motorcycle | 80 | 32 | 40% |
| Sofa | 65 | 14 | 22% |
| Total | 1,781 | 398 | 22% |

We also picked three interesting large objects and reconstructed them using a high-fidelity reconstruction pipeline that performs loop closure detection and global optimization [1]. These three reconstructions are also provided on the web site. We did not apply the high-fidelity reconstruction pipeline to all stationary scans in the dataset because of its high computational requirements.

## Acknowledgements

This project was conducted during the year 2013 while the authors were affiliated with Stanford University. Financial support was provided by the Intel Science and Technology Center for Visual Computing and the Max Planck Center for Visual Computing and Communication.

# References

1. Choi, S., Zhou, Q.Y., Koltun, V.: Robust reconstruction of indoor scenes. In: CVPR (2015) 6
2. Janoch, A., Karayev, S., Jia, Y., Barron, J.T., Fritz, M., Saenko, K., Darrell, T.: A category-level 3-D object dataset: Putting the Kinect to work. In: ICCV Workshops (2011) 1
3. Kerl, C., Sturm, J., Cremers, D.: Robust odometry estimation for RGB-D cameras. In: ICRA (2013) 5, 6
4. Lai, K., Bo, L., Ren, X., Fox, D.: RGB-D object recognition: Features, algorithms, and a large scale benchmark. In: Consumer Depth Cameras for Computer Vision. Springer (2013) 1
5. Newcombe, R.A., Izadi, S., Hilliges, O., Molyneaux, D., Kim, D., Davison, A.J., Kohli, P., Shotton, J., Hodges, S., Fitzgibbon, A.: KinectFusion: Real-time dense surface mapping and tracking. In: ISMAR (2011) 5
6. Singh, A., Sha, J., Narayan, K.S., Achim, T., Abbeel, P.: BigBIRD: A large-scale 3D database of object instances. In: ICRA (2014) 1